\documentclass[twoside,11pt]{article}
\usepackage{jmlr2e}
\usepackage{theorem,url,paralist}
\usepackage{amsmath,latexsym,xspace,float,multirow}
\usepackage{graphics, color,amsmath,verbatim,alltt,subfig}
\def\addr{\small\it}%
\def\email{\hfill\small\sc}%
\def\name{\normalsize\bf}% 

\def\xset{{\mathcal{X}}}
\def\yset{{\mathcal{Y}}}
\def\iSL{{\sc IllinoisSL}\xspace}
\def\by{{\boldsymbol{y}}}

\def\yb{{\boldsymbol{y}}}
\def\bx{{\boldsymbol{x}}}
\def\xb{{\boldsymbol{x}}}
\def\xib{{\boldsymbol{\xi}}}

\def\bw{{\boldsymbol{w}}}
\def\wb{{\boldsymbol{w}}}
\def\bxi{{\boldsymbol{x}_i}}
\def\byi{{\boldsymbol{y}_i}}
\def\bby{\bar{{\boldsymbol{y}}}}

\newcommand*\samethanks[1][\value{footnote}]{\footnotemark[#1]}
\jmlrheading{?}{?}{??-??}{??}{??/??}{}
\ShortHeadings{Illinois-SL: A JAVA-Based Structured Prediction Library}{}
\firstpageno{1}

\begin{document}
\title{IllinoisSL: A JAVA Library for Structured Prediction}
\author{\name Kai-Wei Chang\thanks{
Most of this work was done while the author was at the University of Illinois, supported by DARPA, under 
under agreement number FA8750-13-2-0008. The U.S. Government is authorized to reproduce and distribute reprints for Governmental purposes notwithstanding any copyright notation thereon. The views and conclusions contained herein are those of the authors and should not be interpreted as necessarily representing the official policies or endorsements, either expressed or implied, of DARPA or the U.S. Government.
} \email kw@kwchang.net \\
\addr Microsoft Research New England, MA \AND
\name Shyam Upadhyay \email upadhya3@illinois.edu \\
\addr Department of Computer Science, University of Illinois at 
Urbana-Champaign, IL \AND
\name Ming-Wei Chang \email minchang@microsoft.com \\
\addr Microsoft Research, Redmond, WA \AND
\name Vivek Srikumar\samethanks \email svivek@cs.utah.edu \\
\addr School of Computing at the University of Utah, UT\AND
\name Dan Roth \email danr@illinois.edu \\
\addr Department of Computer Science, University of Illinois at 
Urbana-Champaign, IL 
}

\editor{}

\maketitle

\begin{abstract}%
\iSL is a Java library for learning structured prediction models. It supports structured Support Vector Machines and structured Perceptron.
The library consists of a core learning module and several applications, which can be executed from command-lines.
Documentation is provided to guide users. In Comparison to other structured 
learning libraries, \iSL is efficient, general, and easy to use. 

\end{abstract}

% chapter
%\input{intro}

\section{Introduction} 
\label{sec:intro}
Structured prediction models have been widely used in several fields,
ranging from natural language processing, computer vision, and
bioinformatics. To make structured prediction more accessible to
practitioners, we present \iSL, a Java library for implementing
structured prediction models. Our library supports fast parallelizable
variants of commonly used models like Structured Support Vector
Machines (SSVM) and Structured Perceptron (SP), allowing users to use
multiple cores to train models more efficiently. Experiments on
part-of-speech (POS) tagging show that models implemented in \iSL
achieve the same level of performance as SVM$^{struct}$, a well-known
C++ implementation of Structured SVM, in one-sixth of its training
time. To the best of our knowledge, \iSL is the first fully self-contained
structured learning library in Java. The library is released under
NCSA licence\footnote{http://opensource.org/licenses/NCSA}, providing
freedom for using and modifying the software.

\iSL provides a generic interface for building algorithms to learn from data. 
A developer only needs to define the input and the output structures, and specify the underlying model and inference algorithm (see Sec. 3). 
Then, the parameters of the model can be estimated by the learning algorithms provided by library. 
The generality of our interface allows users to switch seamlessly between several learning algorithms. 

The library and documentation are available
at \url{http://cogcomp.cs.illinois.edu/page/software_view/illinois-sl}.

\section{Structured Prediction Models}
\label{sec:smodel}
% introduce Structured SVM and Structured Perceptron formulation
%\iSL supports Structured SVM and Structured Perceptron. 
This section introduces the notation and briefly describes the
learning algorithms.  We are given a set of training data
$\mathcal{D}=\{\xb_i,\yb_i\}_{i=1}^l$, where instances
$\xb_i \in \xset$ are annotated with structured outputs
$\yb_i \in \yset_i$, and $\yset_i$ is a set of feasible structures for
the $i^{th}$ instance.

Structured SVM \citep{TaskarGuKo04,TJHA05} learns a weight vector
$\wb \in \mathbb{R}^n$ by solving the following optimization problem:
\begin{equation}
  \label{eq:ssvm}
%  \begin{split}
  \min_{\bw, \xib} \quad \frac{1}{2}\bw^T\bw\!+\!C\sum_i \xi^2_i  \quad   s.t.   
  \  \ \bw^T \Phi(\bxi, \byi)\!-\!\bw^T \Phi(\bxi, \by) \geq \Delta(\byi, \by)\!-\!\xi_i, \ \  \forall i, \by \in \yset_i.
%  \end{split}
\end{equation}
where $\Phi(\bx,\by)$ is a feature vector extracted from both input
$\bx$ and output $\by$.  The constraints in \eqref{eq:ssvm} force the
model to assign higher score to the correct output strcture $\byi$
than to others.  $\xi_i$ is a slack variable and we use $L^2$ loss to
penalize the violation in the objective function \eqref{eq:ssvm}. \iSL
supports two algorithms to solve \eqref{eq:ssvm}, a dual coordinate descent method
(DCD)~\citep{CSGR10,ChangYi13} and a parallel DCD algorithm, DEMI-DCD~\citep{ChangSrRo13}.

% Perceptorn
\iSL also provides an implementation of Structured Perceptron~\citep{Collins02}. 
%Averaged Structured Perceptron cycles through the training data. 
At each step, Structured Perceptron updates the model using one
training instance ($\xb_i, \yb_i$) by 
%based on the following update
%rule:
%\begin{equation*}
$\bby \leftarrow \arg\max\nolimits_{\by \in \yset_i} \bw^T \phi(\xb_i, \by), 
	  \bw \leftarrow \bw + \eta (\phi(\xb_i, \yb_i) - 
	\phi(\xb_i, \bby)),$
%\end{equation*}
where $\eta$ is a learning rate. 
%and $\bby$ is the structured output predicted by the current model. 
%The package also includes a variant version, called 
%Averaged Structured Perceptron, which 
%The algorithm maintains a collection of weight vectors encountered during the training.  The final model is a weighted sum of 
%these weight vectors based on . 
Our implementation includes the averaging trick 
introduced in \cite{Daume06}.

\section{IllinoisSL Library}
\label{sec:library}
\begin{table}
% BEGIN RECEIVE ORGTBL thenameofthetable
\begin{tabular}{lllll}
\hline
Task & $\xb$ & $\yb$ & {\sc InfSolver} & {\sc FeatureGenerator} \\
\hline
POS & sentence & tag & Viterbi & Emission and \\
Tagging &  & sequence &  & Transition Features \\
\hline
Dependency & sentence & dependency & Chu-Liu-Edmonds & Edge features \\
Parsing &  & tree &  &  \\
\hline
Cost-Sensitive & document & document & argmax & document features \\
Multiclass &  & category &  &  \\
\hline
\end{tabular}
% END RECEIVE ORGTBL thenameofthetable
\label{table:expample}
\caption{Examples of applications implemented in the library.}
\end{table}

We provide command-line tools to allow users to quickly learn a model for 
problems with common structures, such as linear-chain, ranking, or a 
dependency tree.

The user can also implement a custom structured prediction model through the library interface. We describe how to do the latter below.

\noindent{\bf Library Interface.}
\iSL requires users to implement the following classes:
\begin{compactitem}
\item {\sf IInstance}: the input $\xb$ (e.g., sentence in POS tagging).
\item {\sf IStructure}: the output structure $\yb$ (e.g., tag sequence in POS tagging).
\item {\sf AbstractFeatureGenerator}: contains a function {\sf FeatureGenerator} to extract features  $\phi(\xb,\yb)$ from an example pair $(\xb,\yb)$. 
\item {\sf AbstractInfSolver}: provides a method for solving inference (i.e., 
	$\arg\max_\by \bw^T\phi(\xb_i,\yb)$) and for loss-augmented inference (
	$\arg\max_\by \bw^T\phi(\xb_i,\yb) + \Delta(\yb,\yb_i)$),
and a method for evaluating the loss $\Delta(\yb,\yb_i)$. For example, in POS 
tagging, this class will include implementations of a viterbi decoder and the 
hamming loss, respectively.
\end{compactitem}

\begin{comment}
#+ORGTBL: SEND thenameofthetable orgtbl-to-latex :splice nil :skip 0
|----------------+----------+------------+-----------------+------------------------|
| Task           | $\xb$    | $\yb$      | {\sc InfSolver} | {\sc FeatureGenerator} |
|----------------+----------+------------+-----------------+------------------------|
| POS            | sentence | tag        | Viterbi         | Emission and           |
| Tagging        |          | sequence   |                 | Transition Features    |
|----------------+----------+------------+-----------------+------------------------|
| Dependency     | sentence | dependency | Chu-Liu-Edmonds | Edge features          |
| Parsing        |          | tree       |                 |                        |
|----------------+----------+------------+-----------------+------------------------|
| Cost-Sensitive | document | document   | argmax          | document features      |
| Multiclass     |          | category   |                 |                        |
|----------------+----------+------------+-----------------+------------------------|
\end{comment}
Once these classes are implemented, the user can seamlessly switch between different learning algorithms. 

\noindent{\bf Ready-To-Use Implementations.} 
The \iSL package contains implementations of several common NLP tasks including a sequential tagger, a cost-sensitive mulcticlass classifier, 
and an MST dependency parser.  %based on ~\cite{MPRH05}. 
Table \ref{table:expample} shows the implmentation details of these learners.
%wehre inference was performed using Chu-Liu-Edmonds algorithm~\cite{ChuLi65}.
These implementations provide users with the ability to easily train a model 
for common problems using the command lines, and also serve as examples for 
using the library. The README file provides the details of how to use the 
library.

\noindent{\bf Documentation.}
\iSL comes with detailed documentations, including JAVA API, command-line usage, and a tutorial.
The tutorial provides a step-by-step instructions for building a POS tagger in $~$350 lines of JAVA code. 
Users can post their comments and questions about the package toto \url{illinois-ml-nlp-users@cs.uiuc.edu}.

%\subsection{Command Line Usage}
%#For end users, we provide several applications for users to directly 

\section{Comparison}
\label{sec:comparison}

\begin{figure}[t]
\centering
\begin{tabular}{@{}c@{}c@{}}
\subfloat[Part-of-speech Tagging]{
\includegraphics[width=0.5\linewidth]{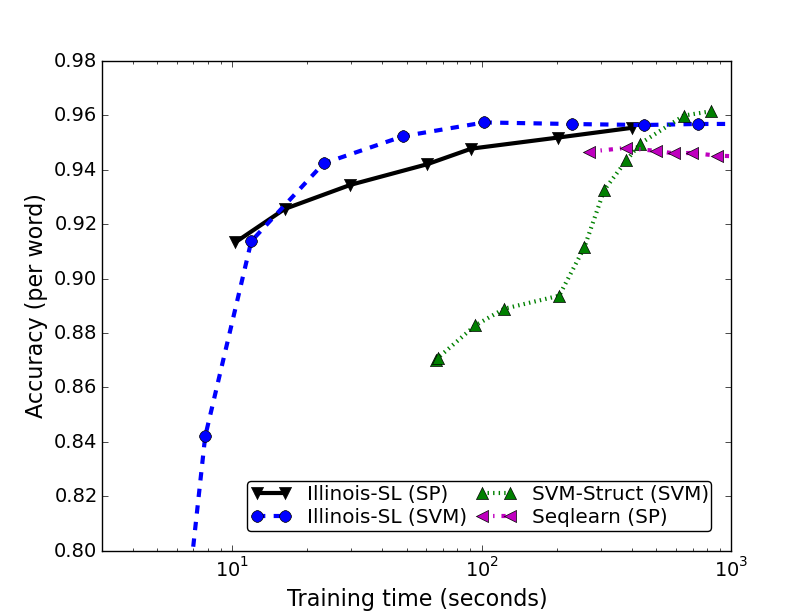}
\label{fig:pos}
}& 
\subfloat[Dependency Parsing]{
\includegraphics[width=0.5\linewidth]{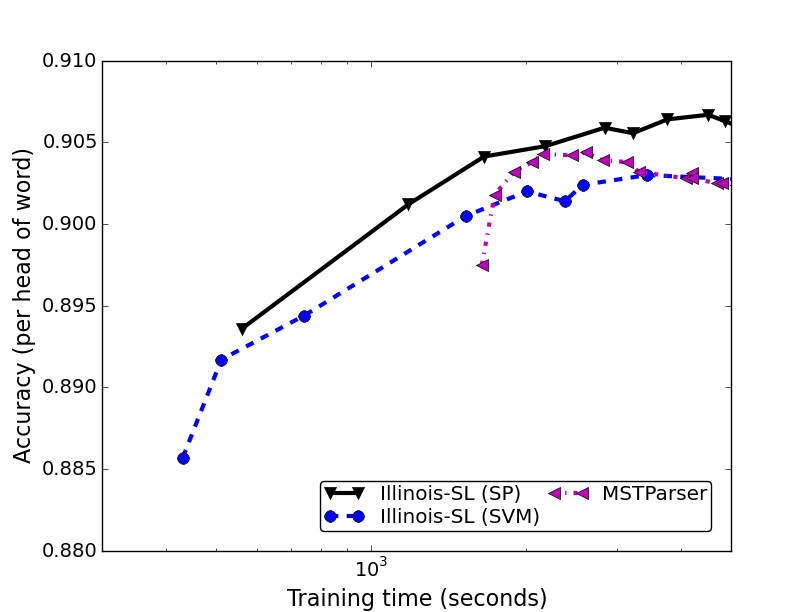}
\label{fig:dp}
}
\end{tabular}
\caption{Accuracy verse training time of two NLP tasks on PTB.}
\label{fig:searchspace}
\end{figure}
To show that \iSL-based implementation of common NLP systems is on par with other structured learning libraries, we compare \iSL with SVM$^{\text{struct}}$\footnote{\url{http://www.cs.cornell.edu/people/tj/svm_light/svm_struct.html}} and Seqlearn\footnote{\url{https://github.com/larsmans/seqlearn}} on a Part-of-speech (POS) tagging problem.\footnote{We do not compare with pyStruct~\citep{MullerBe14} because their package does not support sparse vectors. When representing the features using dense vector, pyStruct suffers from large memory usage and computing time.}
We follow the settings in \cite{ChangSrRo13} and conduct experiments on the English Penn Treebank bank (PTB)~\citep{penn-tree-bank}.
SVM$^{\text{struct}}$ solves an L1-loss structured SVM problem using a cutting-plane method~\citep{JoachimsFiYu09}. Seqlearn implemented a structured Perception algorithm for the sequential tagging problem. For \iSL, we use 16 CPU cores to train the structured SVM model.
Default parameters are used. Figure \ref{fig:pos} shows the accuracy along training time of each model with default parameters.
Despite being a general-purpose package, \iSL is more efficient than 
others\footnote{Note that different learning packages using different training 
objectives.  Therefore, the accuracy performances are slightly different.  }.

% we only used first-order features
We also implemented a minimum spanning tree based dependency parser using \iSL API.
The implementation was done in less than 1000 lines of code, with a few hours of coding effort. 
Figure \ref{fig:dp} shows the performance of our system in accuracy of head words (i.e., unlabeled 
attachment score). \iSL is competitive with MSTParser\footnote{http://www.seas.upenn.edu/~strctlrn/MSTParser/MSTParser.html}, a popular implementation of dependency parser.
%On the PTB dataset, we were able to get a UAS (unlabeled attachment score) of 
%90.3\% with ten passes of structured Perceptron and simple features, which 
%shows that with minimal effort \iSL can support obtaining  competitive results.
%The code for the dependency parser will be released as an example usage along with the package.
%The code that reproduces the experiments is available for download with the package.

%\section{Conclusions}
%\label{sec:conclusion}
%\input{conclusion}

%\bibliographystyle{}
\bibliography{ccg-compact,cited-compact,my}
\end{document}